\documentclass[11pt]{article}

\usepackage[T1]{fontenc}
\usepackage[utf8]{inputenc}
\usepackage{lmodern}
\usepackage[margin=1in]{geometry}
\usepackage{amsmath,amssymb}
\usepackage{graphicx}
\usepackage{booktabs}
\usepackage{tabularx}
\usepackage{array}
\usepackage{microtype}
\usepackage{xurl}
\usepackage[hidelinks]{hyperref}
\usepackage{enumitem}
\usepackage{caption}
\usepackage{float}

\setlist[itemize]{leftmargin=1.6em,itemsep=0.25em,topsep=0.4em}
\newcolumntype{Y}{>{\raggedright\arraybackslash}X}

\begin{document}

\begin{center}
{\LARGE\bfseries Spectral Adapters for Segment Anything Model-based Segmentation of Colorectal Liver Metastases in Computed Tomography\par}
\vspace{1.0em}
{\normalsize
Ramtin Mojtahedi\textsuperscript{a,1,*}, Mohammad Hamghalam\textsuperscript{a,e}, Jacob J. Peoples\textsuperscript{b}, Natalie Gangai\textsuperscript{b}\\
Mithat Gonen\textsuperscript{f}, Yun Shin Chun\textsuperscript{g}, HyunSeon Christine Kang\textsuperscript{h}, Richard K. G. Do\textsuperscript{b}, Amber L. Simpson\textsuperscript{c,d}\par}
\vspace{0.8em}
\begin{minipage}{0.96\textwidth}
\centering\footnotesize
\textsuperscript{a}School of Computing, Queen's University, Kingston, ON, Canada\\
\textsuperscript{b}Department of Radiology, Memorial Sloan Kettering Cancer Center, New York, NY, USA\\
\textsuperscript{c}Department of Radiology and Diagnostic Imaging, University of Alberta, Edmonton, AB, Canada\\
\textsuperscript{d}Alberta Machine Intelligence Institute, Edmonton, AB, Canada\\
\textsuperscript{e}Department of Electrical Engineering, Qazvin Branch, Islamic Azad University, Qazvin, Iran\\
\textsuperscript{f}Department of Epidemiology and Biostatistics, Memorial Sloan Kettering Cancer Center, New York, NY, USA\\
\textsuperscript{g}Department of Surgical Oncology, The University of Texas MD Anderson Cancer Center, Houston, TX, USA\\
\textsuperscript{h}Department of Abdominal Imaging, The University of Texas MD Anderson Cancer Center, Houston, TX, USA\\
\textsuperscript{1}Present address: Toronto General Hospital Research Institute, University Health Network, 200 Elizabeth Street, Toronto, ON M5G 2C4, Canada\\
\textsuperscript{*}Corresponding author: \href{mailto:ramtin.mojtahedi@queensu.ca}{ramtin.mojtahedi@queensu.ca}
\end{minipage}
\vspace{0.8em}

\small\textit{Preprint. Submitted for peer review.}
\end{center}

\begin{abstract}
\textbf{Background and Objectives:} Accurate segmentation of colorectal liver metastases (CRLM) in contrast-enhanced computed tomography (CT) is essential for response assessment, surgical planning, and longitudinal follow-up. Promptable foundation models like the Segment Anything Model (SAM) offer reusable backbones, but full fine-tuning of vision transformers is often infeasible due to GPU and time constraints. We propose two spectral adapter architectures, the Directional Spectral Adapter (DiSECT) and Spectral Instance-Guided Adapter (SiGA), to efficiently adapt SAM for CRLM segmentation in CT.

\textbf{Methods:} We develop SAM-based segmentation models by integrating adapter modules into a frozen SAM Vision Transformer-Base backbone. DiSECT leverages singular value decomposition of frozen weights to constrain residual updates within the leading spectral subspace, while SiGA incorporates global and input-conditioned gating through a multilayer perceptron. We compare DiSECT and SiGA with low-rank adaptation (LoRA), quantized low-rank adaptation (QLoRA), and convolutional adapter (CAD) on 446 contrast-enhanced CT volumes (355 training, 91 testing) using slice-wise experiments across single-point, three-point, bounding-box, and no-prompt regimes.

\textbf{Results:} Compared with standard parameter-efficient fine-tuning techniques, SiGA achieves the highest single-point performance with 0.77 Dice similarity coefficient, 0.69 intersection over union (IoU), and 35.39 mm 95th-percentile Hausdorff distance (HD95). In no-prompt inference, SiGA attains 0.76 Dice, 0.68 IoU, and 46.76 mm HD95, comparable to the nnU-Net baseline (0.758 Dice). DiSECT offers an ultra-lightweight configuration with only 0.14 million trainable parameters.

\textbf{Conclusions:} Spectral adapters enable SAM to deliver competitive CRLM segmentation in CT with minimal trainable parameters, facilitating efficient deployment in resource-constrained clinical environments.
\end{abstract}

\noindent\textbf{Keywords:} Spectral adapters; Medical image segmentation; Liver tumor; Colorectal liver metastases; Segment Anything Model; Parameter-efficient fine-tuning

\section{Introduction}
Liver metastases from colorectal cancer are common and clinically significant, and contrast-enhanced computed tomography (CT) remains the primary imaging modality for detection, staging, and monitoring. Accurate segmentation of colorectal liver metastases (CRLM) supports volumetric tumor burden assessment, radiomics analysis, and treatment response evaluation, but manual delineation is time-consuming and prone to inter-observer variability. Benchmarks such as the Liver Tumor Segmentation (LiTS) challenge show that liver segmentation can reach Dice similarity coefficient (DSC) values around 0.96, whereas tumor segmentation remains harder, with top DSC values between 0.67 and 0.74 \cite{bilic2023}. CRLM-focused studies further highlight difficulties from small lesion size, low contrast, heterogeneous enhancement, scanner differences, and protocol variation \cite{mojtahedi2026self,hamghalam2026,moh2022patch}. Deep convolutional neural networks (CNNs), especially U-Net-based architectures, have advanced liver and tumor segmentation, with nnU-Net serving as a strong self-configuring baseline \cite{isensee2021}. However, nnU-Net and related three-dimensional (3D) U-Net variants remain sensitive to domain shifts and acquisition variability \cite{hamghalam2026,nicoli2025} and usually require separate task-specific networks for different organs, tumor types, scanners, and protocols, making maintenance resource-intensive \cite{mojtahedi2023simsiam}. Recent vision foundation models introduce promptable segmentation: the Segment Anything Model (SAM) segments objects from points, boxes, or masks using a vision-transformer image encoder, prompt encoder, and lightweight mask decoder \cite{kirillov2023}. MedSAM adapts this framework to medical imaging across modalities and anatomical structures \cite{ma2024}, while related medical foundation models combine large-scale pretraining with task-agnostic encoders \cite{liu2026structsam,he2025vista3d}. These models may generalize across segmentation tasks, but full fine-tuning of large transformer backbones is computationally expensive, limiting use where graphics processing unit (GPU) resources and annotated data are constrained \cite{liu2026structsam,he2025vista3d,mojtahedi2023brain}.

SAM can be adapted through zero-shot use, full fine-tuning, prompt tuning, decoder-side fine-tuning, or lightweight encoder adapters \cite{ma2024,wu2025,belton2026,krishnan2026,zhang2026brain}. Medical SAM Adapter (Med-SA), for example, combines lightweight adapters with hyper-prompting to capture domain-specific information with limited parameters \cite{wu2025,jin2024zoo}. However, SAM-based medical segmentation studies suggest that naive fine-tuning and adapter placement can be suboptimal for 3D CT volumes with subtle and heterogeneous lesions, where accuracy and efficiency must be balanced \cite{belton2026,krishnan2026,zhang2026brain}.

Parameter-efficient fine-tuning (PEFT) adapts pretrained models with only a small fraction of trainable parameters while keeping most backbone weights frozen \cite{wang2025survey,chen2023vitadapter,mojtahedi2026peft}. Low-rank adaptation (LoRA) injects low-rank residual matrices into frozen linear layers, while quantized low-rank adaptation (QLoRA) reduces memory by storing frozen weights in low-precision formats \cite{hu2022lora,dettmers2023qlora}. Convolutional adapter (CAD) extends LoRA-style adaptation with depthwise convolutions to capture local spatial structure efficiently \cite{kim2024cad}, and these methods have been applied to vision transformers and SAM-like architectures to reduce training and deployment costs \cite{chen2023vitadapter,mojtahedi2026peft,kim2024cad,mojtahedi2025fewshot}. Recent spectral methods use singular value decomposition (SVD) of pretrained weights, with Spectral Adapter, SVDiff, and weight-decomposed low-rank adaptation (DoRA) constraining updates to compact or directional spectral subspaces \cite{zhang2024spectral,han2023svdiff,liu2024dora}. These approaches suggest that pretrained directional structure can improve learning efficiency and stability, but spectral adaptation has not been systematically investigated for SAM-based CRLM segmentation in CT.

Building on these insights, we propose two spectral adapter architectures for SAM: the Directional Spectral Adapter (DiSECT) and the Spectral Instance-Guided Adapter (SiGA). DiSECT performs residual updates within the leading spectral subspace of frozen transformer weights, while SiGA adds global and input-conditioned gating through a multilayer perceptron for instance-wise routing across lesions with varying size, shape, and contrast. In this work, we evaluate DiSECT and SiGA against established PEFT methods including LoRA, QLoRA, and CAD under multiple prompting strategies, and compare performance with a strong 3D nnU-Net baseline to examine whether lightweight spectral SAM adaptations can match fully trained CNN models while improving the accuracy-efficiency trade-off for biomedical software deployment.

\subsection{Contributions}
The main contributions of this work are:
\begin{itemize}
    \item \textbf{Spectral adapter design for SAM.} We introduce two spectral adapter architectures, DiSECT and SiGA, operating within the singular-vector subspace of frozen transformer weights, with SiGA adding input-conditioned gating for instance-wise routing of spectral directions.
    \item \textbf{Comprehensive CRLM evaluation in CT.} We benchmark DiSECT and SiGA against established PEFT methods, including LoRA, QLoRA, and CAD, under multiple prompting strategies.
    \item \textbf{Comparison with a strong CNN baseline.} We compare adapted SAM models with a 3D nnU-Net baseline for CRLM segmentation, directly evaluating foundation-model adaptation against fully trained task-specific networks.
    \item \textbf{Performance-efficiency analysis.} We analyze segmentation accuracy and computational efficiency across adapters to guide deployment of SAM-based segmentation models in resource-constrained clinical settings.
\end{itemize}

\section{Methods}
\subsection{Problem Formulation and Overall Architecture}
\begin{figure}[H]
    \centering
    \includegraphics[width=\textwidth]{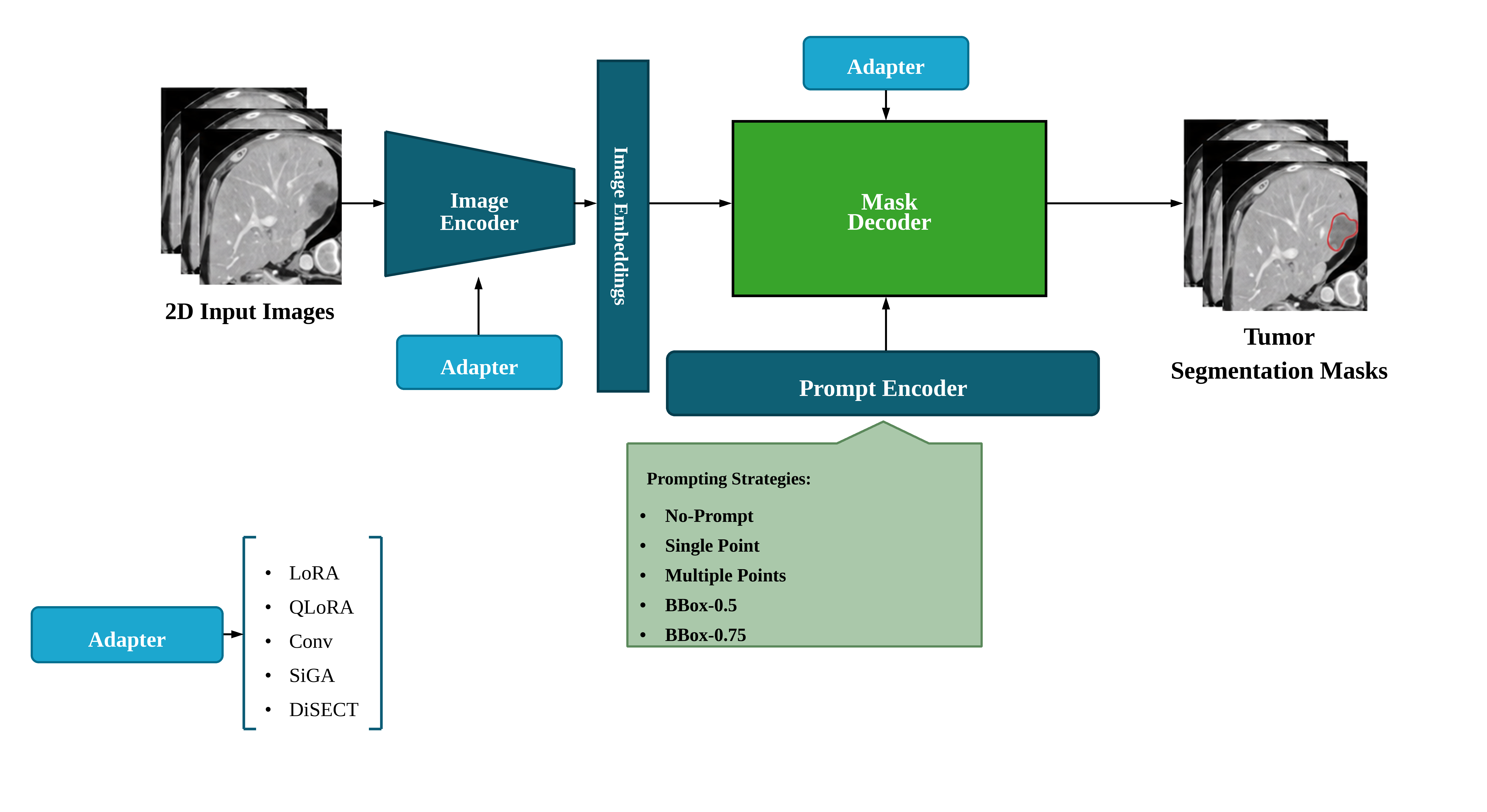}
    \caption{Overview of the proposed pipeline. An adapterized image encoder and a prompt encoder produce embeddings that are fused in the mask decoder to predict tumor masks.}
    \label{fig:pipeline}
\end{figure}

Figure~\ref{fig:pipeline} illustrates the overall segmentation pipeline. Our framework builds on the SAM architecture \cite{kirillov2023}, recently adapted to medical imaging tasks such as CT segmentation \cite{ma2024}. SAM consists of a vision-transformer image encoder, prompt encoder, and mask decoder that fuses image and prompt embeddings to predict segmentation masks. For CRLM segmentation in CT, we add lightweight adapter modules within the image encoder and mask decoder while keeping backbone weights frozen. The main adaptations are spectral adapters operating within the singular-vector subspace of pretrained transformer weights, enabling parameter-efficient task specialization.

Let $x$ denote a $d_{\mathrm{in}}$-dimensional vectorized input to a linear layer with frozen weight $W \in \mathbb{R}^{d_{\mathrm{out}}\times d_{\mathrm{in}}}$ and output $y \in \mathbb{R}^{d_{\mathrm{out}}}$. All adapters keep $W$ frozen and add a parameter-efficient residual branch.

\subsection{Spectral Adapters and Baseline PEFT Methods}
To adapt the SAM backbone for CRLM segmentation, we propose two spectral adapter architectures and compare them with widely used PEFT baselines. The proposed adapters, DiSECT and SiGA, operate within the spectral subspace of frozen transformer weights. For comparison, we also implement three established PEFT methods: LoRA, QLoRA, and CAD. In this context, the adapter rank, $r$, represents the number of retained singular directions for DiSECT and SiGA, whereas for LoRA, QLoRA, and CAD, it represents the width of the low-rank or compact adapter branch. A higher $r$ value gives the adapter more capacity, but it also increases the model's complexity and the number of trainable parameters. The adapter rank for each PEFT method was set to $r=64$ for DiSECT, $r=256$ for SiGA, $r=64$ for LoRA, $r=12$ for QLoRA, and $r=128$ for CAD.

\subsubsection{Spectral Adapters (DiSECT and SiGA)}
For the spectral adapters, we compute an SVD of the frozen weight $W$ and retain the top $r$ singular directions. Let
\[
W \approx U_r \Sigma_r V_r^{\top}
\]
be the rank-$r$ approximation, where $U_r$ and $V_r$ contain the leading left and right singular vectors.

\textbf{DiSECT.} This adapter defines the residual in this spectral subspace:
\begin{equation}
    y = Wx + U_r\left(g \odot \left(V_r^{\top}x\right)\right),
    \label{eq:disect}
\end{equation}
where $g \in [0,1]^r$ is a trainable gate vector and $\odot$ denotes element-wise multiplication \cite{zhang2024spectral,han2023svdiff}. Sparsity in $g$ can restrict the number of active spectral directions, further improving parameter efficiency and stabilizing adaptation.

\textbf{SiGA.} This adapter extends DiSECT with instance-dependent gating. An instance gate $g_{\mathrm{inst}}(x) \in [0,1]^r$ is predicted from the current input features through a lightweight multilayer perceptron and is combined with a global gate $g_{\mathrm{base}} \in [0,1]^r$:
\begin{equation}
    y = Wx + U_r\left(\left(g_{\mathrm{inst}}(x) \odot g_{\mathrm{base}}\right) \odot \left(V_r^{\top}x\right)\right).
    \label{eq:siga}
\end{equation}
This mechanism enables instance-aware routing of spectral directions, allowing the adapter to emphasize or suppress spectral components depending on tumor appearance, size, and contrast.

\subsubsection{Low-Rank Adapters (LoRA and QLoRA)}
\textbf{LoRA.} This adapter adds a low-rank residual:
\begin{equation}
    y = Wx + \alpha BAx,
    \label{eq:lora}
\end{equation}
where $A \in \mathbb{R}^{r\times d_{\mathrm{in}}}$ and $B \in \mathbb{R}^{d_{\mathrm{out}}\times r}$ are trainable matrices, $r \ll \min(d_{\mathrm{in}},d_{\mathrm{out}})$, and $\alpha>0$ is a scalar scale factor \cite{hu2022lora}.

\textbf{QLoRA.} This adapter follows the LoRA formulation but stores the backbone in 4-bit quantized form to reduce memory requirements:
\begin{equation}
    y = W_{\mathrm{4bit}}x + \alpha BAx,
    \label{eq:qlora}
\end{equation}
where $W_{\mathrm{4bit}}$ is a quantized version of $W$ kept in 4-bit precision \cite{dettmers2023qlora}.

\subsubsection{Convolutional Adapter (CAD)}
CAD augments a low-rank residual with depthwise convolutions to capture local structure \cite{kim2024cad}. Given an input $x$,
\begin{equation}
    y = Wx + \eta B\,\mathrm{Mix}(Ax),
    \label{eq:cad}
\end{equation}
where $A$ and $B$ are learned projection layers. In the original convolutional channel adapter, $A$ first maps the feature channels into a compact $r$-dimensional adapter space; for convolutional targets, these projections are implemented as $1\times1$ convolutions. $\mathrm{Mix}(\cdot)$ applies channel-wise multi-scale depthwise convolution in this space, and $B$ projects the result back to the original feature dimension using another $1\times1$ projection. For linear transformer target layers, the same projection-and-mixing idea is implemented using linear projections. $\mathrm{Mix}(\cdot)$ is a depthwise multi-scale dilated convolution applied channel-wise, and $\eta>0$ is a learnable gain.

\subsection{Prompting Strategies}
We evaluate prompt types detailed in Table~\ref{tab:prompts}: a single point, three points, bounding boxes with target intersection over union (IoU) of 0.50 or 0.75, and a no-prompt setting. Point and box prompts are automatically derived from ground-truth tumor masks for controlled evaluation. In the no-prompt setting, no point, box, or mask prompt is provided; predictions use the image input with SAM's empty-prompt embedding. The prompt regimes were examined during development to select a reference training condition. Because single-point prompting gave the highest validation performance while requiring minimal annotation, it was selected for adapter training, model selection, and the main compute comparison; the held-out test analysis is reported under no-prompt inference.

\begin{table}[H]
\centering
\caption{Prompt strategies used in this study.}
\label{tab:prompts}
\small
\begin{tabularx}{\textwidth}{>{\bfseries}p{0.18\textwidth} p{0.27\textwidth} Y}
\toprule
Prompt type & Prompt encoder input & Generation process \\
\midrule
Single point & One $(x,y)$ foreground point & A point placed inside the tumor near the visual center of the selected slice. \\
Three points & Three $(x,y)$ points for foreground or background & One central positive point and two refinement points near tumor boundaries and potential confounders. \\
Box with IoU 0.50 & One rectangular box & A tight ground-truth bounding box resized until the box and the mask reach an IoU value close to 0.50. \\
Box with IoU 0.75 & One rectangular box & A tight ground-truth bounding box resized until the box and the mask reach an IoU value close to 0.75 for finer localization. \\
No-prompt & No prompt input & Automatic mask generation without clicks or boxes at inference time. \\
\bottomrule
\end{tabularx}
\end{table}

\subsection{Reference Fully Trained Baseline}
As a strong fully trained convolutional neural network (CNN) baseline, we use the 3D nnU-Net framework \cite{hamghalam2026,isensee2021}, adapted with residual encoder connections to form a Residual Encoder U-Net (ResEncUNet) \cite{isensee2024revisited}. The network has a seven-stage encoder-decoder structure with feature channels scaling from 32 to 320, using $3\times3\times3$ convolutions, instance normalization, LeakyReLU activations ($\alpha=0.01$), and $2\times2\times2$ strided convolutions for downsampling. It processes $96\times256\times256$ voxel patches and is trained on the training cohort, then evaluated on the same 91 held-out test cases used for the SAM-based models. Final predictions are generated by averaging probabilistic outputs from an ensemble of $N=5$ models and thresholding voxel-wise probabilities at 0.5 to obtain binary tumor masks. Preprocessing follows the standard nnU-Net pipeline with liver-specific adaptations: CT intensities are clipped to the 0.5--99.5 percentile Hounsfield Unit (HU) range ([9, 208] HU), z-score standardized, and resampled to [1.5, 0.8066, 0.8066] mm per voxel to account for multi-institutional slice thickness variability ranging from 0.8 to 7.5 mm.

\subsection{Dataset and Preprocessing}
We utilize a multi-institutional cohort of 446 portal venous phase contrast-enhanced CT volumes from patients with colorectal liver metastases (CRLM), acquired from Memorial Sloan Kettering Cancer Center and University of Texas MD Anderson Cancer Center. Public data from The Cancer Imaging Archive \cite{simpson2024} and retrospective/prospective cohorts across cancer stages and imaging conditions improve diversity and generalizability. Mean voxel spacing is (0.822, 0.822, 4.164) mm; tumor masks were automatically generated and expert-verified. For SAM-based experiments, data are split into 355 training and 91 held-out test cases. We crop the liver and extract 2D axial tumor-positive slices; therefore, no-prompt evaluation uses these slices without point, box, or mask prompts, not full-volume detection or tumor-negative false positives. Windowing over [-150, 250] HU and normalization are applied. Slices are resized to $1024\times1024$ pixels and stored as 2D PNG images with masks, consistent with prior SAM-based methods and reducing memory and computation versus 3D processing.

\subsection{Implementation Details}
We initialize the foundation backbone with public medical adapter weights from the Medical Adapter Zoo \cite{jin2024zoo}, trained via Med-SA \cite{wu2025}; experiments use SAM Vision Transformer-Base (ViT-B) on one NVIDIA A100 GPU (40 GB). During training, only adapters are updated; pretrained weights remain frozen; models differ only in adapter design and QLoRA backbone quantization. Following SAM-style medical adaptation practices \cite{ma2024,wu2025}, we apply standard augmentations and train 2D SAM-adapter models with weighted binary cross-entropy; Dice, IoU, and 95th-percentile Hausdorff distance (HD95) are used for validation/reporting. Models train up to 20 epochs using AdamW, learning rate $1\times10^{-4}$, weight decay 0.01, cosine scheduling, batch size 2, and gradient checkpointing. A validation split is used for model selection and early stopping; the 91 held-out test cases are excluded. Validation occurs every 5 epochs and final epoch; the highest validation-Dice checkpoint under the reference single-point prompt is retained, with early stopping after 5 epochs without improvement.

\subsection{Evaluation Metrics}
Segmentation quality is evaluated using DSC and IoU as overlap metrics and HD95 as a boundary metric \cite{taha2015}. Unless otherwise specified, we report training metrics under the single-point regime and held-out test metrics under the no-prompt regime.

\section{Results}
\subsection{Training Performance and Compute Trade-Offs}
Table~\ref{tab:train} summarizes training performance and computational characteristics under the single-point prompting regime, which achieved the highest performance among the evaluated prompt regimes, including total parameters, trainable parameters, and floating-point operations (FLOPs) per image. All methods share the same SAM backbone and differ only in adapter design. SiGA achieves the best performance, with a DSC of 0.77, IoU of 0.69, and lowest HD95 of 35.39 mm. CAD gives comparable overlap (DSC = 0.76) but with the highest computational cost, while LoRA reaches a DSC of 0.75 with moderate trainable parameters and good throughput. QLoRA slightly lowers overlap (DSC = 0.74) but achieves the fastest throughput due to backbone quantization. DiSECT is the most parameter-efficient method, with only 0.14 million trainable parameters, or 0.14\% of total parameters after rounding, but has lower overlap accuracy and larger boundary error, reflecting the trade-off of extreme parameter efficiency.

\begin{table}[H]
\centering
\caption{Training performance under the single-point prompting regime with compute metrics. Floating-point operations are reported in giga FLOPs per image. Latency and throughput are measured during validation. Trainable (\%) is computed as $100\times$ trainable parameters/total parameters.}
\label{tab:train}
\scriptsize
\resizebox{\textwidth}{!}{%
\begin{tabular}{lrrrrrrrrr}
\toprule
Adapter & DSC & IoU & HD95 (mm) & FLOPs (G) & Params (M) & Trainable (M) & Trainable (\%) & Latency (ms) & Throughput (img/s) \\
\midrule
SiGA   & \textbf{0.77} & \textbf{0.69} & \textbf{35.39} & 865.92  & 121.26 & 22.18 & 18.29 & 289.15 & 3.46 \\
CAD    & 0.76 & 0.67 & 41.04 & 1263.69 & 124.05 & 28.51 & 22.98 & 134.83 & 7.42 \\
LoRA   & 0.75 & 0.66 & 41.18 & 882.87 & 113.92 & 9.54 & 8.37 & 121.25 & 8.25 \\
QLoRA  & 0.74 & 0.65 & 45.66 & 882.87 & 113.92 & 1.79 & 1.57 & \textbf{115.20} & \textbf{8.68} \\
DiSECT & 0.70 & 0.62 & 52.55 & \textbf{770.01} & \textbf{99.22} & \textbf{0.14} & \textbf{0.14} & 158.65 & 6.30 \\
\bottomrule
\end{tabular}}
\end{table}

\subsection{Test Performance Under No-Prompt Setting}
Table~\ref{tab:test} reports segmentation performance on the held-out test cohort of 91 cases under the no-prompt inference setting, which reflects segmentation without user-specified point or box prompts. SiGA again achieves the highest overlap performance, attaining the best DSC and IoU on the test set while maintaining competitive boundary accuracy. CAD narrows the gap in overlap metrics but retains a relatively large HD95. LoRA performs similarly to CAD in overlap but shows inferior boundary accuracy. QLoRA, despite its reduced memory footprint from backbone quantization, matches SiGA in boundary accuracy but falls behind in overlap. DiSECT yields the lowest overlap and the largest boundary error, consistent with its emphasis on extreme parameter efficiency.

\begin{table}[H]
\centering
\caption{Segmentation performance on the held-out test cohort of 91 cases under no-prompt inference.}
\label{tab:test}
\begin{tabular}{lrrr}
\toprule
Adapter & DSC & IoU & HD95 (mm) \\
\midrule
SiGA   & \textbf{0.76} & \textbf{0.68} & 46.76 \\
CAD    & 0.73 & 0.65 & 48.06 \\
LoRA   & 0.74 & 0.65 & 57.88 \\
QLoRA  & 0.70 & 0.61 & \textbf{46.30} \\
DiSECT & 0.70 & 0.61 & 60.11 \\
\bottomrule
\end{tabular}
\end{table}

\subsection{Contextual Comparison with nnU-Net}
To contextualize the performance of adapterized SAM-based models, Table~\ref{tab:nnunet} compares tumor DSC with the fully trained nnU-Net baseline. The SAM adapter values are the same no-prompt results measured on tumor-positive 2D slices in Table~\ref{tab:test}, while the nnU-Net value is the average tumor DSC from the 3D volume-based ensemble. Because IoU and HD95 were not computed under an identical 3D protocol for the baseline, DSC is used as the common metric. The nnU-Net baseline achieves 0.758 DSC, while SiGA achieves 0.76 DSC. The methods also differ in inference settings and training paradigm: nnU-Net is trained and evaluated on fully supervised 3D volumes, whereas adapterized SAM models perform slice-wise 2D segmentation with frozen backbones. Despite this difference, the results indicate that a parameter-efficient spectral adapter can approach the performance of a strong fully trained 3D baseline, while preserving promptability and the ability to be reused for other segmentation tasks.

\begin{table}[H]
\centering
\caption{Comparison of tumor Dice similarity coefficient between adapterized SAM variants under no-prompt inference and the nnU-Net baseline.}
\label{tab:nnunet}
\begin{tabular}{lr}
\toprule
Method & Tumor DSC \\
\midrule
SiGA (SAM-based) & \textbf{0.76} \\
nnU-Net (three-dimensional CNN) & 0.758 \\
CAD (SAM-based) & 0.73 \\
LoRA (SAM-based) & 0.74 \\
QLoRA (SAM-based) & 0.70 \\
DiSECT (SAM-based) & 0.70 \\
\bottomrule
\end{tabular}
\end{table}

\section{Discussion}
Among the evaluated parameter-efficient adapters integrated into the SAM-based backbone, the proposed SiGA consistently delivers the highest segmentation accuracy in training and testing. Under single-point prompting during training, SiGA achieves a DSC of 0.77, outperforming CAD, LoRA, QLoRA, and DiSECT, and on the held-out no-prompt test set reaches a DSC of 0.76 with favorable IoU and HD95 values. This advantage can be attributed to its spectral dual-gating mechanism, which combines a global gate capturing layer-wise spectral importance with an instance gate adapting spectral directions to the current input, aligning residual updates with informative spectral modes while allowing case-specific modulation. From an efficiency point of view, LoRA and QLoRA provide attractive trade-offs: LoRA maintains moderate trainable parameters and good throughput with only a small DSC drop relative to SiGA, while QLoRA is appealing under memory constraints because it quantizes the frozen backbone, achieves the fastest throughput, and retains comparable boundary accuracy, although with lower overlap. CAD reduces the performance gap to SiGA but demands more FLOPs, while DiSECT is the most parameter-efficient option, though its lower overlap and higher boundary error suggest that purely global spectral gating may not fully capture colorectal liver metastasis heterogeneity. These behaviors reflect design trade-offs, allowing adapters to be selected based on accuracy, memory, or compute constraints. Compared with nnU-Net, which remains a strong task-specific baseline with a tumor DSC of 0.758, the adapterized SAM-based SiGA model achieves comparable accuracy with a 0.76 DSC while updating only a fraction of the backbone parameters and retaining the ability to be re-prompted or adapted to other tasks. For centers already using nnU-Net, SAM-based adapters can therefore serve as a complementary strategy for reusing a single foundation backbone across several organs and pathologies.

\subsection{Impact of Spectral Structure and Prompting}
The spectral adapters DiSECT and SiGA are inspired by work on spectral fine-tuning \cite{zhang2024spectral,han2023svdiff} and directional parameter-efficient fine-tuning \cite{liu2024dora}. Our results suggest that aligning residual updates with leading singular directions is helpful but not sufficient on its own. The instance-wise gating in SiGA appears vital in the presence of high inter-case variability. Prompting strategies also strongly influence performance. Single-point prompts provide strong tumor-focused supervision during training, but no-prompt testing is more realistic for high-throughput clinical workflows. The no-prompt results show that adapterized SAM-based models can produce accurate automatic segmentations without prompt interaction, supporting the design choice of combining global spectral structure with instance-specific modulation in SiGA.

\subsection{Limitations and Future Work}
This study has several limitations. First, we use 2D slices from 3D volumes, discarding through-plane context that may improve tumor detection and boundary refinement; extending spectral adapters to 3D encoder-decoder architectures or emerging 3D SAM variants is a natural next step. Second, experiments focus on portal-venous-phase colorectal liver metastases, so cross-site generalization, multi-phase performance, and robustness to protocol changes remain to be studied. Third, adapter ranks, spectral dimensionality, and gating-network capacity were not exhaustively explored, and joint optimization may yield further gains. Future work can examine hybrid architectures combining 3D CNNs such as nnU-Net with adapterized foundation backbones, incorporate multi-phase CT or multi-modal inputs with advanced prompts, and extend spectral adapters to federated or edge deployments where parameter-efficient fine-tuning is especially attractive.

\section{Conclusion}
We presented an evaluation of spectral and low-rank PEFT strategies for a SAM-based foundation model for colorectal liver metastasis segmentation in contrast-enhanced CT. Integrating LoRA, QLoRA, CAD, DiSECT, and SiGA into a frozen SAM backbone, we showed that SiGA achieved the best accuracy, reaching 0.76 test DSC under realistic no-prompt inference. This is competitive with a fully trained 3D nnU-Net baseline (0.758 DSC) while using fewer trainable parameters and preserving promptable flexibility. When accuracy is the priority, SiGA is preferred; when memory or latency dominate, QLoRA and LoRA offer high-throughput alternatives. CAD is suitable when added compute is acceptable, while DiSECT remains viable for extreme parameter-efficient settings. Overall, spectral instance guidance is key for adapting foundation models to heterogeneous medical imaging tasks, supporting adapterized SAM-based models as practical, efficient, reusable segmentation engines for liver tumors and clinical applications.

\section*{CRediT Author Statement}
Conceptualization: R.M., M.H., M.G., A.L.S.; Methodology: R.M., M.H., M.G., A.L.S.; Formal analysis and Visualization: R.M.; Data curation: N.G., Y.S.C., H.C.K., M.G., R.K.G.D.; Funding acquisition and Resources: R.K.G.D., A.L.S.; Supervision: A.L.S.; Writing -- original draft: R.M., M.H.; Writing -- review \& editing: all authors.

\section*{Funding}
NIH/NCI grant R01CA233888.

\section*{Ethics Approval and Consent}
De-identified human CT images/masks were used under Queen's University HSREB approval for DMED-2441-21 (TRAQ \#6031742; renewed Oct. 30, 2025, valid to Dec. 8, 2026). Consent/waiver requirements followed the approved protocol; no recruitment, contact, intervention, identifiable-information collection, or animal studies occurred.

\section*{Competing Interests}
A.L.S. and R.K.G.D. report NCI/NIH financial support; others report no known competing interests.

\section*{Data/Code Availability}
Institutional data are restricted; TCIA data are available from \cite{simpson2024}; code will be released upon acceptance.

\section*{Generative AI Statement}
ChatGPT (OpenAI) was used only for limited English editing; the authors reviewed the content and take responsibility.

\end{document}